\documentclass[final,1p,times]{elsarticle}

\usepackage{booktabs}
\usepackage{lipsum}
\makeatletter
\def\ps@pprintTitle{%
 \let\@oddhead\@empty
 \let\@evenhead\@empty
 \def\@oddfoot{}%
 \let\@evenfoot\@oddfoot}
\makeatother

\usepackage{graphicx}
\usepackage{amssymb}
\usepackage{amsmath}
\usepackage{commath}
\usepackage{mathtools}

\DeclareMathOperator*{\argmax}{arg\,max}
\DeclareMathOperator*{\argmin}{arg\,min}

\usepackage{float}

\begin{document}

\begin{frontmatter}

\title{Dynamic Time Warp Convolutional Networks}
\author{Yaniv Shulman}
\address{yaniv@aleph-zero.info}

\begin{abstract}
Where dealing with temporal sequences it is fair to assume that the same kind of deformations that motivated the development of the Dynamic Time Warp algorithm could be relevant also in the calculation of the dot product ("convolution") in a 1-D convolution layer. In this work a method is proposed for aligning the convolution filter and the input where they are locally out of phase utilising an algorithm similar to the Dynamic Time Warp. The proposed method enables embedding a non-parametric warping of temporal sequences for increasing similarity directly in deep networks and can expand on the generalisation capabilities and the capacity of standard 1-D convolution layer where local sequential deformations are present in the input. Experimental results demonstrate the proposed method exceeds or matches the standard 1-D convolution layer in terms of the maximum accuracy achieved on a number of time series classification tasks. In addition the impact of different hyperparameters settings is investigated given different datasets and the results support the conclusions of previous work done in relation to the choice of DTW parameter values. The proposed layer can be freely integrated with other typical layers to compose deep artificial neural networks of an arbitrary architecture that are trained using standard stochastic gradient descent.
\end{abstract}
\end{frontmatter}

\section{Introduction}
\label{S:Introduction}

\emph{Convolutional Neural Network} (CNN) is a family of machine learning algorithms that have gained importance in recent years and displayed demonstrable success tackling difficult problems in areas of research such as machine vision and time series classification. An in-depth review of \emph{Artificial Neural Networks} (ANN) and CNNs is available in \cite{DBLP:journals/corr/Schmidhuber14, Goodfellow-et-al-2016}. The fundamental component of a CNN is the \emph{convolution layer}, a layer that is capable of learning jointly with other components a set of parameters that enables the entire network to achieve its learning task. The convolution layer can readily combine with other typical ANN layers such as a \emph{fully connected layer} and a \emph{pooling layer} to provide a rich end-to-end approach to develop deep learning models that can be trained using gradient decent methods. Deep learning methods such as CNNs alleviate in most cases the need to perform elaborate pre-processing of the inputs and in particular alleviate the need to hand engineer features which typically require expert domain knowledge, is time-consuming and error prone. \newline

\emph{Dynamic Time Warping} (DTW) \cite{1163055} is an algorithm that enables aligning two temporal sequences under assumptions of local deformations in time such as local accelerations and decelerations and has been used extensively since its inception in solving problems in many domains where temporal sequences are involved. Utilising DTW and k-NN for classification has consistently obtained state-of-the-art results in many reported temporal sequences classification tasks \cite{WangYO16, ratanamahatana2004everything, IsmailFawaz2019}. \newline

In this work a subset of the family of CNNs is considered where the input is limited to either a univariate or multivariate temporal sequence such as time series data or any data that can be converted into a temporal sequence, these CNNs are referred to as \emph{1-D CNNs}. A novel \emph{DTW convolution layer} (DTW-CNN) is proposed that combines an algorithm similar to the DTW and the standard 1-D convolution layer to improve on the generalisation capabilities of the standard convolution layer to local deformations in time. The proposed DTW-CNN is evaluated on a number of time series classification tasks to demonstrate the effects of replacing a standard 1-D convolution layer with a DTW convolution layer. In summary the key contributions of this work are:
\begin{enumerate}
\item The proposal of the DTW-CNN layer, a novel 1-D convolution layer that is designed to be invariant to local deformations.
\item Embedding a non-parametric warping aspect of temporal sequences similarity directly in deep networks.
\end{enumerate}

\section{Preliminaries}
\label{S:Preliminaries}
In this section a review of the Dynamic Time Warping algorithm is given to set the notation and naming conventions used in subsequent sections. \newline \newline
Dynamic Time Warping \cite{1163055} is a method to calculate a mapping between two temporal sequences such that the misalignment between the two sequences is minimised. Let $Q$ be a sequence of length $N \in \mathbb{N} $ and $C$ be  a sequence of length $M \in \mathbb{N}$ such that $Q = q_1, q_2, \ldots, q_n, \ldots, q_N$ and $C = c_1, c_2, \ldots, c_m, \ldots, c_M$. The two sequences can be aligned using DTW as follows. First construct a local distance matrix $D \in \mathbb{R}^{N \times M} \coloneqq d(q_n, c_m) = \norm{q_n - c_m}$. A warping path is an ordered mapping $P=\{p_1, p_2, \ldots, p_K\}$ with $p_k \in \{1, \ldots, N\} \times \{1, \ldots, M\}$ that associates an element $q_n \in Q$ with $c_m \in C$ under the following constraints:

\begin{enumerate}
\item Boundary condition: $p_1 = (1, 1)$ and $p_K = (N, M)$.
\item Monotonicity condition: $n_1 \leq n_2 \leq \ldots \leq n_K$ and $m_1 \leq m_2 \leq \ldots \leq m_K$.
\item Continuity condition: $n_k - n_{k-1} \leq 1$ and $m_k - m_{k-1} \leq 1$.
\end{enumerate}

The mapping that minimises the misalignment between the two sequences is a path through the matrix $D$ that minimises the warping cost:

\begin{equation}
\label{eq:dtw_dist1}
P^* = \argmin_{P \in \mathcal{P}}{\{Cost (P)\}}
\end{equation}
where
\begin{equation}
\label{eq:dtw_dist1}
Cost (P) = \sum_{k=1}^{K} D_{p_k}
\end{equation}

where $D_{p_k}$ is the matrix $D$ entry indexed by the $k^{th}$ element of a warping path $P$ and $K$ is the length of the path $P$. The path $P^*$ is referred to as the \emph{optimal path}. \newline

DTW utilises a dynamic programming equation to find the optimal path with complexity $O(NM)$ or lower using heuristics or approximations \cite{prune_dtw, Salvador:2007:TAD:1367985.1367993, lucky_dtw, msdtw} . In the most straightforward approach also referred to as unconstrained DTW it does so by calculating a cumulative cost matrix $G \in \mathbb{R}^{N \times M}$ as follows:
\begin{enumerate}
\item First row: $G(1, m) = \sum_{k=1}^{m} d(q_1, c_m), m \in [1, M]$
\item First column: $G(n, 1) = \sum_{k=1}^{n} d(q_n, c_1), n \in [1, N]$
\item All other elements: \newline $G(n, m) = \argmin{\{G(n - 1,m - 1), G(n - 1, m), G(n, m - 1) \}} + d(q_n, c_m)\\ n \in [2, N],  m \in [2, M]$ \newline
\end{enumerate}

This formulation of the DTW algorithm has a number of drawbacks:
\begin{enumerate}
\item Paths can map very short sections of one sequence to long sections of the other resulting in unrealistic and in many cases undesirable deformations.
\item The number of paths grows exponentially with $N$ and $M$.
\item Paths lengths are at most  $N + M - 1$ which represents a significant deviation from the length of the original sequences.
\end{enumerate}

To alleviate these issues and to reduce computational complexity additional constraints and modifications had been proposed to the DTW algorithm:

\begin{enumerate}
\item Adjustment window condition: imposes a global constraint on the distance that the path can take from the main diagonal of the matrix $D$. Two common formulations exist: the \emph{Sakoe-Chiba Band} such that $\abs{n-m} < r$ where $r \in \mathbb{N}$ is referred to as the \emph{window length} \cite{1163055}; and the \emph{Itakura Parallelogram} in \cite{itakura1}.
\item Slope condition: a limitation on the "gradient" of the path to mitigate local unrealistic alignment between short sequences to long sequences. This is achieved by putting a limit on the number of consecutive steps that can be taken in a direction before a step in the other direction must be taken \cite{1163055}.
\end{enumerate}

\section{Problem Description}
Where dealing with temporal sequences it is fair to assume that the same kind of deformations that motivated the development of the DTW algorithm could be relevant also in the calculation of the dot product ("convolution") in the first convolution layer, and possibly also in subsequent layers. Considering a particular filter in a 1-D convolution layer, it is modified during the training process so that it has a maximised response when a certain feature is present in a section of the input. However when local deformations are present in the input resulting in a local phase mismatch the response of the filter could decrease. Therefore it is desirable for convolution layers to be invariant to such deformations and by doing so possibly improve the generalisation capabilities of a CNN for temporal sequence analysis.

\section{Proposed Method}
\label{S:Proposed Method}
Let $\mathbf{w}=(w_1, w_2, \ldots, w_N)$ be a filter of length $N$ and $\mathbf{x}=(x_t, x_{t+1}, \ldots, x_{t + N - 1})$ be a section of 1-D input starting at index $t$  of equal length. Let the \emph{product matrix} $D \in \mathbb{R}^{N \times N} \coloneqq d(w_i, x_j) = w_i x_j$. Having calculated the product matrix $D$, it is possible using an approach similar to DTW to find the optimal path $P^*$ in $D$ under the same conditions (Boundary, Monotonicity, Continuity) as defined in section \ref{S:Preliminaries}. Formally we would like to calculate:

\begin{equation}
\label{eq:optimal_path1}
P^* = \argmax_{P \in \mathcal{P}}{\{Cost (P)\}}
\end{equation}
where
\begin{equation}
\label{eq:proposed_cost1}
Cost (P) = \sum_{k=1}^{K} D_{p_ k}u(p_k)
\end{equation}

where $D_{p_k}$ is the matrix $D$ entry indexed by the $k^{th}$ element of a warping path $P$, $K$ is the length of the path $P$ and $u(p_k)$ is a normalising function $u(\cdot): \{1, \ldots, N\} \times \{1, \ldots, N\} \mapsto \mathbb{R}$. 
As opposed to DTW, the path $P^*$ is the optimal path that maximises the filter response to the input signal under the deformations that can be realised within the limits of the DTW conditions. In many cases increasing the number of terms in \eqref{eq:proposed_cost1} will lead to larger sums and therefore longer paths are more likely to emerge as $P^*$. Therefore the normalising function is required to normalise the cost in relation to the number of terms in the path. \newline

A basic normalising function that can be suggested assigns an equal weight to each of the terms and thus maintains the full alignment of the two signals however adjusts for the path's length:

\begin{equation}
\label{eq:u_of_x_normalising_3}
u(\cdot) = \frac{1}{\abs{P}}
\end{equation}

where $\abs{P}$ is the length of the path $P$. Therefore equation \eqref{eq:proposed_cost1} can be rewritten as:
\begin{equation}
\label{eq:symmetric_cost}
Cost (P) = \frac{1}{\abs{P}} \sum_{k=1}^{K} D_{p_ k}
\end{equation}
This weighting scheme is referred to as \emph{symmetric} and effectively calculates the mean response of the filter under different alignments as defined by each path $P$.  \newline

The cost of a path $P$ as defined in equation \eqref{eq:proposed_cost1} is by definition a sum of products where all elements of $\mathbf{x}$ and $\mathbf{w}$ are included. By rearranging the terms and grouping by elements of $\mathbf{w}$ the calculation of the cost of path $P$ can be rewritten as:
\begin{equation}
\label{eq:rewritten_cost1}
Cost (P) = \sum_{i=1}^{N} w_i \sum_{j \in X_{wi}} x_j u(i, j)
\end{equation}

where $X_{wi}=\{j|x_j\, is \, multiplied \, by \, w_i \, in \, eq \, \eqref{eq:proposed_cost1}\}$. One possible way to define $u(i,j)$ is as follows:

\begin{equation}
\label{eq:u_of_x_normalising_1}
u(i,j) = 
\begin{cases}
\frac{1}{\abs{X_{wi}}} & (i,j) \in P \\
0 & (i,j) \notin P
\end{cases}
\end{equation}

where $\abs{X_{wi}}$ is the number of elements in $X_{wi}$. Therefore under the joint formulation in equations \eqref{eq:rewritten_cost1} and \eqref {eq:u_of_x_normalising_1} equation \eqref{eq:proposed_cost1} can be rewritten as:
\begin{equation}
\label{eq:x_onto_w_cost}
Cost (P) = \sum_{i=1}^{N} \frac{w_i}{\abs{X_{wi}}} \sum_{j \in X_{wi}} x_j
\end{equation}

The cost formulation in equation \eqref{eq:x_onto_w_cost} can be seen as a weighted alignment of $\mathbf{x}$ onto $\mathbf{w}$ and is referred to \emph{x onto w} in subsequent sections. There is however a natural alternative rearrangement of equation \eqref{eq:proposed_cost1} that can be seen as a weighted alignment of \emph{$\mathbf{w}$ onto $\mathbf{x}$} and is defined by regrouping the cost of the path $P$ by elements of $\mathbf{x}$ with an alternative normalising function:

\begin{equation}
\label{eq:rewritten_cost2}
Cost (P) = \sum_{j=1}^{N} x_j \sum_{i \in W_{xj}} w_i u(i,j)
\end{equation}

where $W_{xj}=\{i|w_i\, is \, multiplied \, by \, x_j \, in \, eq \, \eqref{eq:proposed_cost1}\}$. In a similar fashion one possible way to define $u(i,j)$ is as such:

\begin{equation}\
\label{eq:u_of_x_normalising_2}
u(i,j) = 
\begin{cases}
\frac{1}{\abs{W_{xj}}} & (i,j) \in P \\
0 & (i,j) \notin P
\end{cases}
\end{equation}

Therefore under the joint formulation in equations \eqref{eq:rewritten_cost2} and \eqref {eq:u_of_x_normalising_2} equation \eqref{eq:proposed_cost1} can be rewritten as:
\begin{equation}
\label{eq:w_onto_x_cost}
Cost (P) = \sum_{j=1}^{N} \frac{x_j}{\abs{W_{xj}}} \sum_{j \in W_{xj}} x_j
\end{equation}

All three formulations of the normalised cost in equations \eqref{eq:symmetric_cost}, \eqref{eq:x_onto_w_cost} and \eqref{eq:w_onto_x_cost} can be expressed in matrix notation as:

\begin{equation}
\label{eq:cost1_matrix_normalising_1}
Cost (P) = \mathbf{w} U\mathbf{x} '
\end{equation}
Where in this case $U$ is fully determined by the elements of $P$ and the normalising function $u(\cdot)$. Note that equation \eqref{eq:cost1_matrix_normalising_1} is differentiable in respect 
to $\mathbf{w}$. \\

For example, given the path $P$:
$$
P = \{(0,0), (0,1), (0,2), (1,2), (1,3), (2,3), (2,4), (3,5), (4,6), (5,6), (6,6)\} \\
$$
these are the matrices $U$ corresponding for each of the formulations:
$$
U_{x\,\,onto\,\,w}=
\begin{bmatrix}
\frac{1}{3} & \frac{1}{3} & \frac{1}{3} & 0 & 0 & 0 & 0 \\
0 & 0 & \frac{1}{2} & \frac{1}{2} & 0 & 0 & 0 \\
0 & 0 & 0 &\frac{1}{2} & \frac{1}{2} & 0 & 0 \\
0 & 0 & 0 & 0 & 0 & 1 & 0 \\
0 & 0 & 0 & 0 & 0 & 0 & 1 \\
0 & 0 & 0 & 0 & 0 & 0 & 1 \\
0 & 0 & 0 & 0 & 0 & 0 & 1 \\
\end{bmatrix}
$$
$$
U_{w\,\,onto\,\,x}=
\begin{bmatrix}
1 & 1 & \frac{1}{2} & 0 & 0 & 0 & 0 \\
0 & 0 & \frac{1}{2} & \frac{1}{2} & 0 & 0 & 0 \\
0 & 0 & 0 & \frac{1}{2} & 1 & 0 & 0 \\
0 & 0 & 0 & 0 & 0 & 1 & 0 \\
0 & 0 & 0 & 0 & 0 & 0 & \frac{1}{3} \\
0 & 0 & 0 & 0 & 0 & 0 & \frac{1}{3} \\ 
0 & 0 & 0 & 0 & 0 & 0 & \frac{1}{3} \\
\end{bmatrix}
$$
$$
U_{symmetric}=
\begin{bmatrix}
\frac{1}{11} & \frac{1}{11} & \frac{1}{11} & 0 & 0 & 0 & 0 \\
0 & 0 & \frac{1}{11} & \frac{1}{11} & 0 & 0 & 0 \\
0 & 0 & 0 & \frac{1}{11} & \frac{1}{11} & 0 & 0 \\
0 & 0 & 0 & 0 & 0 & \frac{1}{11} & 0 \\
0 & 0 & 0 & 0 & 0 & 0 & \frac{1}{11} \\
0 & 0 & 0 & 0 & 0 & 0 & \frac{1}{11} \\ 
0 & 0 & 0 & 0 & 0 & 0 & \frac{1}{11} \\
\end{bmatrix}
$$

Note that in all cases the non zero elements of the matrix $U$ are the path $P$'s indices and moreover in $U_{x\,\,onto\,\,w}$ the rows sum to 1, in $U_{w\,\,onto\,\,x}$ the columns sum to 1 and in $U_{symmetric}$ the entire matrix sums to 1. \\

Regardless of a formulation choice the output of the filter is then defined as:
\begin{equation}
\label{eq:filter_output}
y = f(\mathbf{w} U^* \mathbf{x}' + b)
\end{equation}

where $f$ is a non-linear activation, $U^*$ is the matrix corresponding with the optimal path $P^*$ and $b$ is an optional bias scaler. The path following the main diagonal of the matrix $D$ corresponds with $U = I_N$ the identity matrix and thus reduces the output of equation \eqref{eq:filter_output} to the standard convolution layer filter output. To generate a feature map equation \eqref{eq:filter_output} is applied to multiple filters and input sections by repeating the process for each filter/input section combination. Note there is no prevention to use a parameter sharing regime to generate the feature maps. Lastly the computation in practice of the output of the filter in equation \eqref{eq:filter_output} can be done more efficiently than in the naive form described above in \eqref{eq:filter_output} since $U^*$ is relatively sparse.

\subsection{Discussion}

\subsubsection{Hyperparameters}
The proposed DTW-CNN layer expands on the hyperparameters that need to be configured for a standard 1-D CNN layer such as the number of filters, the filter size and the stride. The added hyperparameters are the choice of alignment between the signals (x onto w, w onto x, or symmetric), adjustment window condition size, and the slope. According to \cite{Dau:2018:ODT:3238306.3238325} the optimal adjustment window condition size parameter $r$ in relation to accuracy is data dependent, and in some rare cases does not even matter. Lower values of $r$ are desirable since in the very least will result is less calculations to compute the DTW, also intuitively excessive values of $r$ can result in extreme and unrealistic deformations. Historically the speech recognition community used 10\% warping constraint, however in \cite{ratanamahatana2004everything} it is claimed that even 10\% is too large for real world data.

Experimental results are given in section \ref{S:Experimental Results} to demonstrate the impact of the choice of alignment and the adjustment window condition size on the accuracy for some classification tasks. The slope condition impact was not looked into in this work.

\subsubsection{Architecture}
Technically there are no limitations on where a DTW-CNN layer can be used along the computation graph however considering the class of deformations it is designed to deal with it seems reasonable that it would be useful as the first layer immediately following the input layer so that it can compute activations that are invariant to local out of phase deformations in the input. In addition it seems reasonable to separate multivariate signals and feed each channel through a separate DTW-CNN since it is feasible for different channels to have local deformations that are different in nature and/or location. The individual feature maps output by each DTW-CNN can be concatenated into a single feature map or kept separate before feeding into the next layer(s), this approach is somewhat similar to the MC-DCNN architecture suggested in \cite{Zheng2014TimeSC}.

\subsubsection{Training}
There are a number of considerations in regard to the training of networks that contain DTW-CNN layers. The first is the impact of the transformation $U^*$ on the gradients calculated during the backpropagation algorithm. This can be reasoned about by rewriting equation \eqref{eq:filter_output} as $y = f(\mathbf{w} (U^* \mathbf{x}') + b)$ which follows from the commutativity property of matrix multiplication. This highlights that the gradient updates is similar to a standard CNN in relation to the weights however the inputs to the filter are the product $U^*\mathbf{x}'$, a deterministic transformation of the input dependant on $\mathbf{w}$ and $\mathbf{x}$. Given $\mathbf{w}$ is fixed during the forward and backward propagation $\mathbf{w}$ can be considered as a parameter of the Path cost function \eqref{eq:proposed_cost1} thus making the cost function and the choice of $U^*$ a function of $\mathbf{x}$ only during each and every training iteration. This is consistent with the calculation during inference where $\mathbf{w}$ is constant and therefore the Path cost function \eqref{eq:proposed_cost1} is a function of $\mathbf{x}$ only. 

To give another explanation that may be more insightful it is possible to consider an alternative brute force implementation of equation \eqref{eq:filter_output} by a static computation graph. Let $\mathcal{U}$ be the set of all matrices corresponding with the set of all possible paths ${\mathcal{P}}$ given a choice of normalising function $u(\cdot)$ and a filter of length $N$. Instead of calculating $U^*$ by the dynamic programming method a graph is constructed where the input $\mathbf{x}$ is multiplied by each and every $U \in \mathcal{U}$ followed by a $max(\cdot)$ operation such that $y=max(\{\mathbf{w} U\mathbf{x}' + b|U \in \mathcal{U}\})$. The $max(\cdot)$ operation performs as a gradient flow selection channelling the backpropagation of gradients through the branch of the graph corresponding with the maximal value of all such products.  It is also clear that all operations are differential with respect to $\mathbf{w}$ therefore enabling standard backpropagation of gradients.

\section{Related work}
There exists a huge body of work related to CNN and to the DTW algorithm relating to both theoretical aspects and to solving specific problems using these methods. Due to the volume of work it is impossible to provide a complete review of papers related to the CNN and to the DTW and instead the reader is referred to a number of summary resources and to the references included therein. A review of deep learning for time series classification is done in \cite{IsmailFawaz2019}. A comparison of different methods for time series classification including a few CNN architectures is available in \cite{WangYO16}. An overview of deep learning in neural networks in given in \cite{DBLP:journals/corr/Schmidhuber14}. In \cite{gamboa2017deep} a review is available of various deep learning techniques for time series analysis. A comparative benchmark and a review of various methods for time series classification is available in \cite{DBLP:journals/corr/BagnallBLL16}. A survey of similarity measures for time series is given in \cite{Serr__2014} which determined that DTW and Time Warped Edit Distance (TWED) are the two best performing measures.
  
In \cite{cui2016multiscale} it is suggested to augment inputs by down-sampling and applying lowpass filters to the inputs before using multiple independent filters for each generated input. With this approach the augmentation operation is fixed, and furthermore the approach has the drawback of increasing the number of parameters in the model. Invariant Scattering Convolution Networks \cite{DBLP:journals/corr/abs-1203-1513} computes a translation invariant image representation which is stable to deformations utilising wavelet transform convolutions with non-linear modulus and averaging operators. NeuralWarp jointly learns a deep embedding of the time series with a warping neural network that learns to align values in the latent space \cite{grabocka2018neuralwarp} however the warping function is soft and with no constraints imposed which is likely to result in unrealistic warping. Insights in regard to the optimal warping window size are provided in \cite{Dau:2018:ODT:3238306.3238325} where it determined that the optimal window size is both data and dataset size dependent. In addition \cite{Dau:2018:ODT:3238306.3238325} proposes methods to learn the warping window size and demonstrates that by setting the warping window size correctly most or all the improvement gap of the more sophisticated methods proposed in recent years is matched.

\section{Experimental Results}
\label{S:Experimental Results}
\subsection{Datasets}
To evaluate different aspects of the proposed method four time series datasets are used:

\begin{enumerate}
\item The Photometric LSST Astronomical Time-series Classification Challenge \footnote{http://www.timeseriesclassification.com/description.php?Dataset=LSST} \cite{181000000:online, The:UEA:UCR}. The test and train sets are merged, shuffled and split.
\item Crop \footnote{http://www.timeseriesclassification.com/description.php?Dataset=Crop} \cite{wei_tan_2017, The:UEA:UCR}. The test and train sets are swapped.
\item InsectWingbeatSound \footnote{http://www.timeseriesclassification.com/description.php?Dataset=InsectWingbeatSound} \cite{DBLP:journals/corr/ChenWBMK14, The:UEA:UCR}. The test and train sets are merged, shuffled and split, only observations with exactly seven channels is included.
\item Time Series Land Cover Classification Challenge \footnote{https://sites.google.com/site/dinoienco/tiselc}.
\end{enumerate}

\begin{table}[h!]
\centering
\begin{tabular}{l c c c c c} 
\toprule
Dataset & classes & channels & length & train & test \\
\midrule
LSST & 24 & 6 & 36 & 3447 & 1478 \\ 
Crop & 24 & 1 & 46 & 16800 & 7200 \\
Insect & 10 & 7 & 199 & 18454 & 7910 \\
TiSeLaC & 9 & 10 & 23 & 81714 & 17973 \\
\bottomrule
\end{tabular}
\caption{\label{tab:all means results}Summary of datasets, channels refers to the number of attributes in in each time step; length is the overall length of each example; train is the number of examples in the train set; test is the number of examples in the test set, these are used for evaluating the accuracy.}
\end{table}

\newpage

\subsection{Architecture}
All experiments shared the following basic architecture and settings:
\begin{enumerate}
\item Input layer.
\item For each channel in the input either a standard convolution layer or a DTW convolution layer is created that takes as input the entire sequence for that particular channel, configured with a learned bias and Relu activation.
\item Pooling layer configured with pool size of 2 and a stride of 2.
\item Convolution layer operating on the entire input volume, configured with 128 filters of size 5, stride of 1, a learned bias and Relu activation.
\item Fully connected layer with 512 units, followed by batch normalisation, Relu activation and dropout layer with 0.5 drop probability.
\item Fully connected layer with 256 units, followed by batch normalisation, Relu activation and dropout layer with 0.5 drop probability.
\item Fully connected logits layer.
\item Softmax cross-entropy layer.
\end{enumerate}

For optimisation Adam \cite{Adam} with the default parameter values is employed. Note that there was no attempt to find an optimal architecture or hyperparameter settings for the experiments but merely to measure the relative accuracy for different hyperparameter settings and between the DTW-CNN to a standard CNN under this basic network configuration. Lastly unless explicitly stated when DTW-CNN is used it is used during both training and inference.

\subsection{Hyperparameter experimentation}
\subsubsection{Methodology}
The methodology used for evaluating impact of different hyperparameters settings is straightforward and is based on comparing the accuracy of a classification task. The evaluation controls for the other factors by fixing the data, the architecture and the other hyperparameters when running each experiment. To evaluate the impact of different hyperparameters settings the classification accuracy on the test set is calculated and recorded every time after a number of epochs are fed through the network for training.

\begin{figure}[t]
\centering
\includegraphics[width=0.6\textwidth]{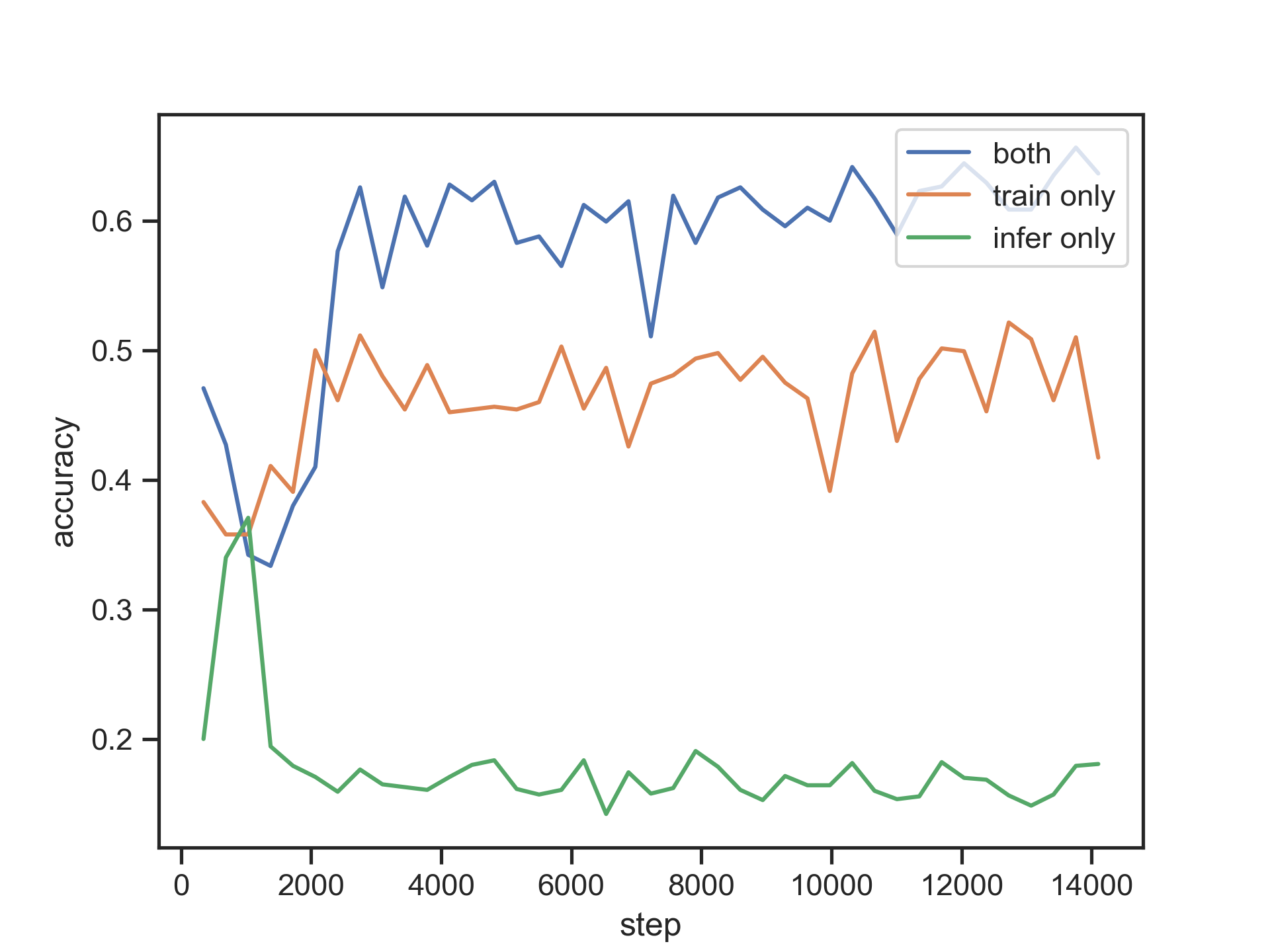}
\caption{Accuracy for the LSST dataset obtained by applying the $U^*$ transform to the inputs during both training and inference, only during training or only during inference.}
\label{fig:apply_u}
\end{figure}

\subsubsection{When to apply $U^*$?}
An interesting aspect to consider is whether it is best to apply the transformation $U^*$ only in training, only in inference, or in both. If it is sufficient to apply only in inference with good results then the computational complexity during training is substantially reduced. The rationale for applying only during inference is that having learnt the filters' weights of a standard CNN in relation to the data it might still be useful to align the inputs against the weights during inference to account for local deformations that may be present in the input, and by doing so improve accuracy. Three different configurations in relation to the application of the DTW transform $U^*$ are trialled:

\begin{enumerate}
\item Both during training and during inference.
\item Only during training.
\item Only during inference.
\end{enumerate}

For this purpose dataset 1 (LSST) was used. The first DTW-CNN layer (for each channel) was configured to have 8 filters of size 7 with a stride of 1, the warping window $r$ was set to 1, a symmetric weighting as per equation \eqref{eq:symmetric_cost}, and a mini-batch size of 100 was used in all experiments. The feature maps output by the individual DTW-CNN layers were concatenated before pooling. The results indicate that the maximum accuracy is achieved by a large margin when the DTW transform $U^*$ is applied both in training and inference, and the worst result was obtained when done only in inference as illustrated in figure \ref{fig:apply_u}. These results can be explained by hypothesising that when $U^*$ is applied only during inference it is equivalent to attempting inference on data that contain substantial local deformations that were not commonly present during training and thus demonstrate the adverse affect it has on a standard CNN as the discriminative model learns $p(y|x,w)$ whereas $p(y|U^*x, w)$  seems to be substantially different for the tested data resulting in substantially reduced accuracy. However when applied only during training the warping maps similar inputs that have local deformations being equivalent to some sort of "pre-clustering" of the inputs which in turn restricts learning to a subspace of the actual training data.

\subsubsection{Adjustment window condition}
To evaluate the impact of modifying the adjustment window condition hyperparameter $r$ the accuracy obtained for different values of $r$ was measured while holding all other hyperparameters fixed for the datasets 1 (LSST) and 2 (Crop). The warping window $r$ was set to values ranging from 1 to 4, a mapping of onto w as per equation \eqref{eq:x_onto_w_cost}, and a mini-batch size of 50 was used in all experiments. The results demonstrate that the network trained successfully for all $r$ values.

\begin{figure}[t]
\centering
\includegraphics[width=0.49\textwidth]{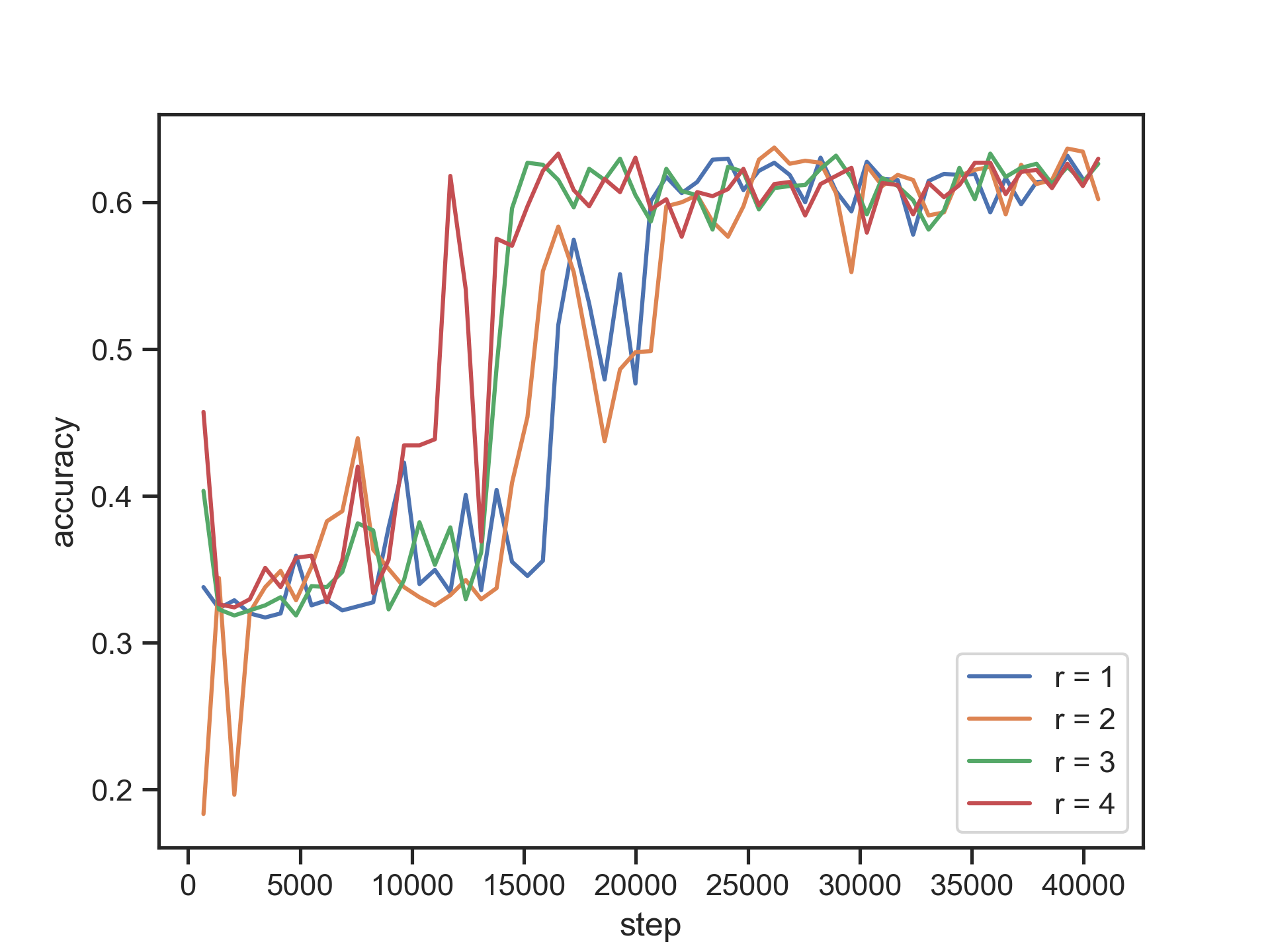}
\includegraphics[width=0.49\textwidth]{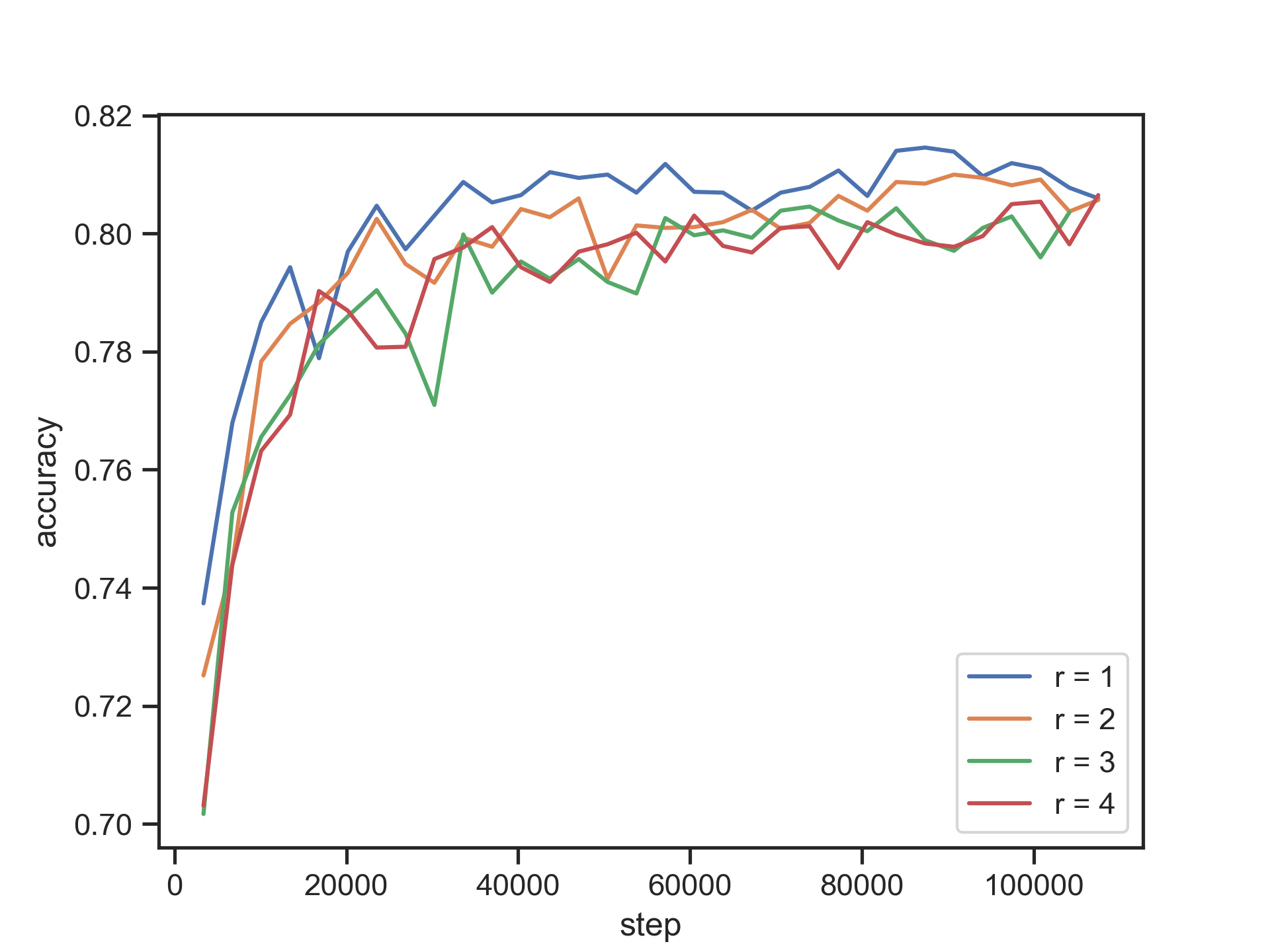}
\caption{Accuracy for the LSST dataset (left) and the Crop dataset (right) for the adjustment window condition hyperparameter $r$ values ranging from 1 to 4.}
\label{fig:crop_r}
\end{figure}

For the LSST dataset the first DTW-CNN layer was configured to have 8 filters of size 7 with a stride of 1 for each channel. In this setting all values of $r$ gave about the same accuracy. It is notable that larger values of $r$ resulted in less steps required for training by roughly 30\% until convergence as illustrated in figure \ref{fig:crop_r}. 

For the Crop dataset the first DTW-CNN layer was configured to have 64 filters of size 7 with a stride of 2. The results show that the network trained successfully for all $r$ values and that $r = 1$ gave the best accuracy by a relatively small margin as illustrated in figure \ref{fig:crop_r}. 

The results demonstrate that $r$ is data dependent and whilst overall accuracy may not increase by increasing $r$, in the context of the DTW-CNN larger values of $r$ may contribute to faster convergence.

\subsubsection{Normalising choice}
In section \ref{S:Proposed Method} three approaches to normalise the path cost are described:
\begin{enumerate}
\item symmetric, equation \eqref{eq:symmetric_cost}.
\item x onto w, equation \eqref{eq:x_onto_w_cost}.
\item w onto x, equation \eqref{eq:w_onto_x_cost}.
\end{enumerate}
To evaluate the effect of modifying the normalising choice $u(\cdot)$ the accuracy obtained for different normalising methods was measured while holding all other hyperparameters fixed for the datasets 1 (LSST) and 3 (Insect).
For the LSST dataset the first DTW-CNN layer was configured to have 32 filters of size 7 with a stride of 1 for each channel and with the warping window $r=1$.
For the Insect dataset the first DTW-CNN layer was configured to have 10 filters of size 7 with a stride of 5 for each channel and with the warping window $r=2$.

\begin{figure}[h]
\centering
\includegraphics[width=0.49\textwidth]{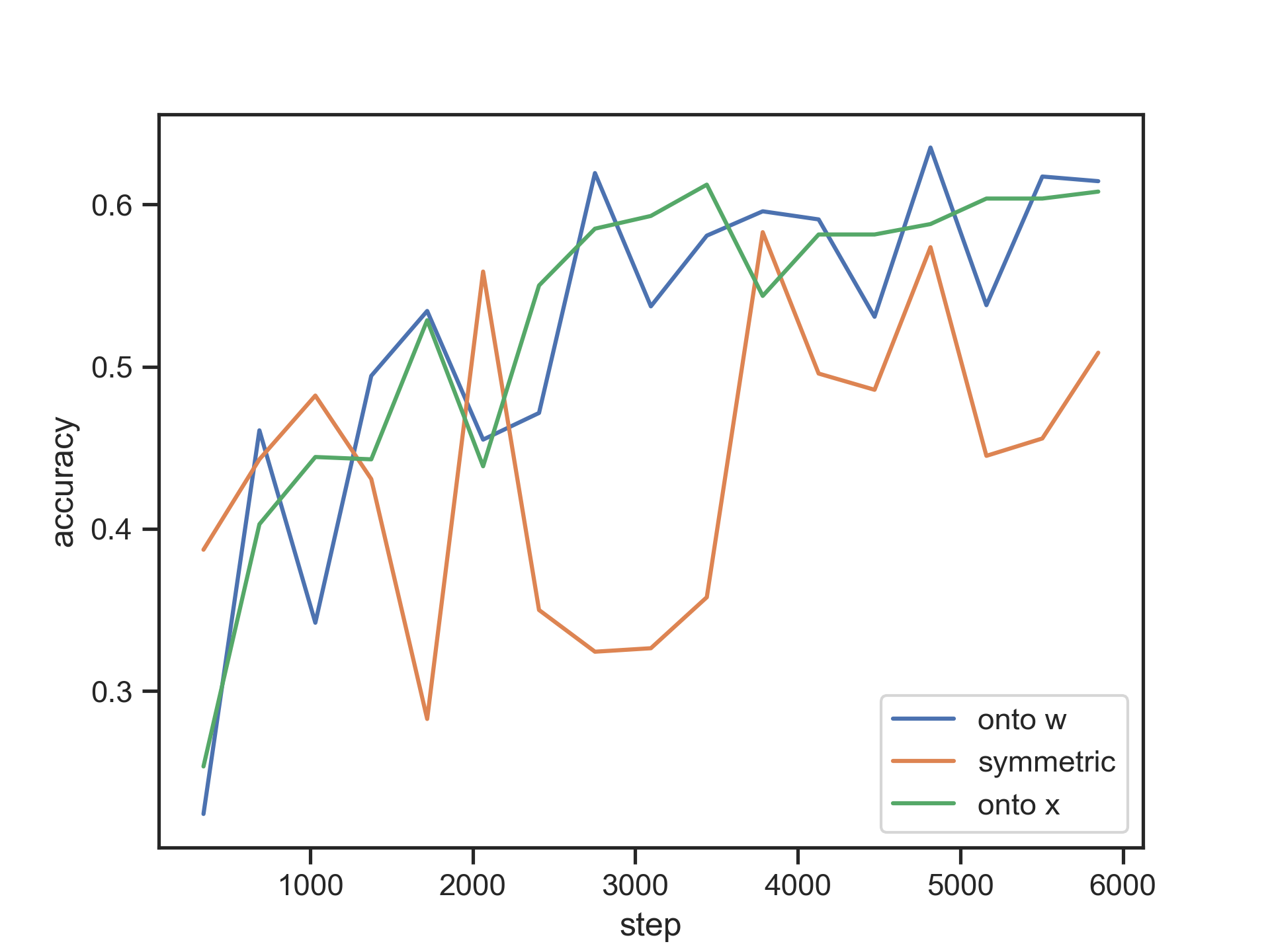}
\includegraphics[width=0.49\textwidth]{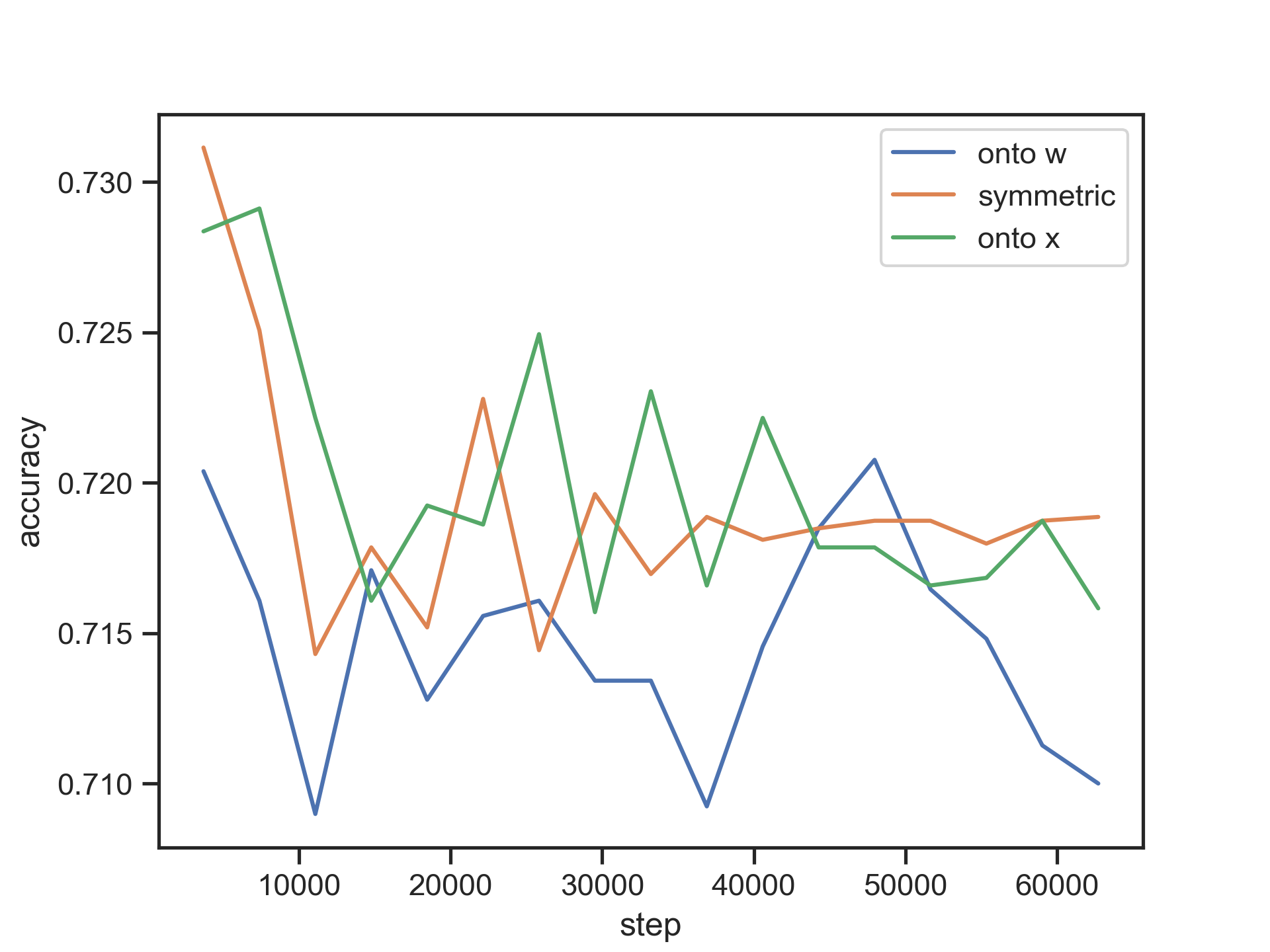}
\caption{Accuracy for the LSST dataset (left) and the Insect dataset (right) given different choices of the normalising function $u(\cdot)$ during training.}
\label{fig:norm}
\end{figure}

The results indicate that the optimal choice of path cost normalisation is dependent on the dataset and possibly other factors as illustrated in figure \ref{fig:norm}. 

\subsection{Comparison against a standard CNN}
\subsubsection{Methodology}
The methodology used for evaluating the effectiveness of the proposed DTW-CNN layer is straightforward and is based on comparing the accuracy of a classification task when using a standard 1-D convolution layer against when using the proposed DTW-CNN layer. The evaluation controls for the other factors by fixing the data, the architecture and the hyperparameters when running each experiment. To be explicit note that when DTW-CNN is used it is used during both training and inference.

\subsubsection{Metrics}
\label{S:Metrics}
The classification accuracy on the test set is calculated and recorded every time after a number of epochs are fed through the network. Three metrics are calculated to estimate the performance of the DTW-CNN layer against the standard 1-D convolution layer:

\begin{enumerate}
\item Mean accuracy $s_1$ to $s_2$, the mean accuracy reported on the test set from training iteration $s_1$ to iteration  $s_2$.
\item Std accuracy $s_1$ to $s_2$, the standard deviation of accuracy reported on the test set from training iteration $s_1$ to iteration $s_2$.
\item Max accuracy $s_1$ to $s_2$,the maximum accuracy reported on the test set from training iteration $s_1$ to iteration $s_2$.
\end{enumerate}

\subsubsection{Results}
The following tables summarise the results obtained for the metrics described in \ref{S:Metrics} when replacing the first CNN layer with a DTW-CNN layer. The results demonstrate that the classification accuracy for the DTW-CNN layer is mostly better, and never significantly worse than a CNN in the tested settings where the magnitude of improvement in accuracy varies across datasets. The variance in accuracy as training progresses when employing the DTW-CNN layer is comparable to the standard CNN layer indicating that standard optimisers do not experience significant problems to converge with the DTW-CNN layer included in the graph. Moreover merely increasing the number of filters and/or their length does not bridge the gap in results between the two type of layers implying that the DTW-CNN is capable of learning representations that fundamentally go beyond the capacity of the standard CNN layer given data with certain characteristics. 

\begin{table}[h!]
\centering
\caption{Comparison of results for dataset 1 (LSST) where the first convolution layer was configured with 8 filters of size 5, stride of 1, $r=1$, x onto w and a mini-batch size of 100.}
\begin{tabular}{*7c}
\toprule
Iterations &  \multicolumn{3}{c}{DTW-CNN} & \multicolumn{3}{c}{CNN}\\
\midrule
{} & mean & std & max & mean & std & max  \\
344 - 4128 & \textbf{0.5127} & 0.094 & \textbf{0.635} & 0.4894 & \textbf{0.0776} & 0.6021 \\
4472 - 8256 & \textbf{0.5925} & 0.0413 & \textbf{0.6293} & 0.554 & \textbf{0.0327} & 0.5921 \\
8600 - 12384 & \textbf{0.6129} & \textbf{0.0208} & \textbf{0.6479} & 0.5599 & 0.0537 & 0.6171 \\
12728 - 16512 & \textbf{0.6014} & \textbf{0.0270} & \textbf{0.6293}  & 0.548 & 0.0513 & 0.6043 \\
16856 - 20640 & \textbf{0.5969} & \textbf{0.0364} & \textbf{0.6236} & 0.5547 & 0.043 & 0.6079 \\
\bottomrule
\end{tabular}
\end{table}

\begin{table}[h!]
\centering
\caption{Comparison of results for dataset 1 (LSST) where the first convolution layer was configured with 128 filters of size 7, stride of 1, $r=1$, x onto w and a mini-batch size of 100.}
\begin{tabular}{*7c}
\toprule
Iterations &  \multicolumn{3}{c}{DTW-CNN} & \multicolumn{3}{c}{CNN}\\
\midrule
{} & mean & std & max & mean & std & max  \\
344 - 4128 & \textbf{0.5405} & 0.0885 & \textbf{0.6336} & 0.5081 & \textbf{0.0737} & 0.5979 \\
4472 - 8256 & \textbf{0.6039} & \textbf{0.0299} & \textbf{0.6343} & 0.529 & 0.0667 & 0.6114 \\
8600 - 12384 & \textbf{0.6057} & 0.0399 & \textbf{0.64} & 0.5665 & \textbf{0.0201} & 0.5979 \\
12728 - 16512 & \textbf{0.6184} & \textbf{0.0201} & \textbf{0.6357}  & 0.5434 & 0.0562 & 0.6007 \\
16856 - 20640 & \textbf{0.617} & \textbf{0.0126} & \textbf{0.635} & 0.5660 & 0.0401 & 0.6171 \\
\bottomrule
\end{tabular}
\end{table}

\begin{table}[h!]
\centering
\caption{Comparison of results for dataset 3 (InsectWingbeatSound) where the first convolution layer was configured with 8 filters of size 5, stride of 2, $r=1$, x onto w and a mini-batch size of 50.}
\begin{tabular}{*7c}
\toprule
Iterations &  \multicolumn{3}{c}{DTW-CNN} & \multicolumn{3}{c}{CNN}\\
\midrule
{} & mean & std & max & mean & std & max  \\
1845 - 18450 &  \textbf{0.7463} & 0.0066 &  \textbf{0.7538} & 0.7353 &  \textbf{0.0065} & 0.7485 \\
20295 - 36900 &  \textbf{0.7524} &  \textbf{0.0021} &  \textbf{0.7572} & 0.7381 & 0.0034 & 0.7432 \\
38745 - 55350 &  \textbf{0.7527} & 0.0051 &  \textbf{0.758} & 0.74 &  \textbf{0.005} & 0.7458 \\
57195 - 73800 &  \textbf{0.7501} & 0.0036 &  \textbf{0.7562}  & 0.7402 &  \textbf{0.0033} & 0.7465 \\
75645 - 92250 &  \textbf{0.7524} & 0.0026 &  \textbf{0.7565} & 0.7426 &  \textbf{0.0025} & 0.7471 \\
\bottomrule
\end{tabular}
\end{table}

\begin{table}[h!]
\centering
\caption{Comparison of results for dataset 3 (InsectWingbeatSound) where the first convolution layer was configured with 64 filters of size 7, stride of 1, $r=1$, x onto w and a mini-batch size of 50.}
\begin{tabular}{*7c}
\toprule
Iterations &  \multicolumn{3}{c}{DTW-CNN} & \multicolumn{3}{c}{CNN}\\
\midrule
{} & mean & std & max & mean & std & max  \\
1845 - 18450 &  \textbf{0.7499} & \textbf{0.0034} &  \textbf{0.7557} & 0.7422 &  0.0044 & 0.751 \\
20295 - 36900 &  \textbf{0.7511} &  0.0039 &  \textbf{0.7585} & 0.7381 & \textbf{0.0019} & 0.7457 \\
38745 - 55350 &  \textbf{0.7525} & \textbf{0.0028} &  \textbf{0.7568} & 0.7414 &  0.0038 & 0.7449 \\
57195 - 73800 &  \textbf{0.7524} & 0.0026 &  \textbf{0.7546}  & 0.7446 &  \textbf{0.0019} & 0.7473 \\
75645 - 92250 &  \textbf{0.7533} & \textbf{0.0034} &  \textbf{0.7585} & 0.7425 &  0.0058 & 0.7503 \\
\bottomrule
\end{tabular}
\end{table}

\begin{table}[h!]
\centering
\caption{Comparison of results for dataset 4 (Satellite) where the first convolution layer was configured with 16 filters of size 7, stride of 1, $r=1$, x onto w and a mini-batch size of 50.}
\begin{tabular}{*7c}
\toprule
Iterations &  \multicolumn{3}{c}{DTW-CNN} & \multicolumn{3}{c}{CNN}\\
\midrule
{} & mean & std & max & mean & std & max  \\
3268 - 52288 &  \textbf{0.9195} & \textbf{0.01869} &  \textbf{0.9343} & 0.919 & 0.02072  & 0.9342 \\
55556 - 104576 &  0.9356 &  0.00199 &  0.9391 & \textbf{0.9372} & \textbf{0.00179} & \textbf{0.9399} \\
107844 - 153596 &  0.9394 & \textbf{0.00214} &  \textbf{0.9441} & \textbf{0.9401} &  0.00227 & 0.9435 \\
156864 - 205884 &  0.9417 & \textbf{0.00167} &  0.9441  & \textbf{0.9425} &  0.00171 & \textbf{0.9454} \\
209152 - 258172 &  \textbf{0.9435} & \textbf{0.00117} &  \textbf{0.9454} & \textbf{0.9435} &  0.00122 & \textbf{0.9454} \\
\bottomrule
\end{tabular}
\end{table}
 
\newpage

\section{Conclusion}
In this article a novel DTW-CNN layer is proposed that combines a 1-D convolution layer with an algorithm similar to the DTW to align the convolution kernel against the inputs. The DTW-CNN layer enables embedding a non-parametric warping of temporal sequences for increasing similarity directly in deep networks. Combining the similarity warping with learned kernel weights results in an overall warping path that minimises the learning task, therefore it can expand on the generalisation capabilities and the capacity of standard 1-D convolution layer where local sequential deformations are present in the input. The results demonstrate that the DTW-CNN exceeds or matches the standard CNN layer in terms of the maximum accuracy achieved on a number of time series classification tasks. In addition the impact of different hyperparameters settings is demonstrated given different datasets and the results support the conclusions of previous work done in relation to the choice of DTW parameter values.

\section{Acknowledgement}
The author wishes to thank Patrick Peursum for his insightful remarks throughout this research.

\bibliographystyle{abbrv}
\bibliography{refs}

\end{document}